\documentclass[runningheads]{llncs}
\usepackage[T1]{fontenc}
\usepackage{graphicx}
\usepackage{xcolor}
\usepackage{multirow}
\usepackage{ulem}

\newcommand{\code}[1]{\texttt{\small#1}}

\usepackage{makecell} 
\usepackage[hidelinks]{hyperref}

\makeatletter
\renewcommand\section{\@startsection{section}{1}{\z@}%
  {8pt plus 4pt minus 0pt}
  {8pt plus 2pt minus 0pt}
  {\normalfont\large\bfseries}}

\renewcommand\subsection{\@startsection{subsection}{2}{\z@}%
  {8pt plus 4pt minus 0pt}
  {4pt plus 2pt minus 0pt}
  {\normalfont\normalsize\bfseries}}
\makeatother

\begin{document}
\title{TutorGym: A Testbed for Evaluating AI Agents as Tutors and Students}

%

\author{
  Daniel Weitekamp$^*$\orcidID{0000-0003-0079-8000} \and \\ 
  Momin N. Siddiqui$^*$\orcidID{0000-0003-1874-7789} \and \\ 
  Christopher J. MacLellan\orcidID{0000-0003-3084-5189}
}
\institute{Georgia Institute of Technology, Atlanta, GA 30332, USA \\
\email{\{weitekamp, msiddiqui66, cmaclell\}@gatech.edu} \\
$(^*)$ Equal contribution 
}

\authorrunning{D. Weitekamp et al.}

\maketitle

\begin{abstract}



Recent improvements in large language model (LLM) performance on academic benchmarks, such as MATH and GSM8K, have emboldened their use as standalone tutors and as simulations of human learning. However, these new applications require more than evaluations of final solution generation. We introduce {\bf TutorGym}\footnote{{\bf Website:} \textcolor{blue}{\url{https://tutorgym.ai}}; {\bf GitHub:} \textcolor{blue}{\href{https://github.com/Teachable-AI-Lab/tutor\_gym}{Teachable-AI-Lab/tutor\_gym}}}
to evaluate these applications more directly. TutorGym is a standard interface for testing artificial intelligence (AI) agents within existing intelligent tutoring systems (ITS) that have been tested and refined in classroom studies, including Cognitive Tutors (CTAT), Apprentice Tutors, and OATutors. TutorGym is more than a simple problem-solution benchmark, it situates AI agents within the interactive interfaces of existing ITSs. At each step of problem-solving, AI agents are asked what they would do as a \textit{tutor} or as a \textit{learner}. As tutors, AI agents are prompted to provide tutoring support---such as generating examples, hints, and step-level correctness feedback---which can be evaluated directly against the adaptive step-by-step support provided by existing ITSs. As students, agents directly learn from ITS instruction, and their mistakes and learning trajectories can be compared to student data. TutorGym establishes a common framework for training and evaluating diverse AI agents, including LLMs, computational models of learning, and reinforcement learning agents, within a growing suite of learning environments. Currently, TutorGym includes 223 different tutor domains.
In an initial evaluation, we find that current LLMs are poor at tutoring---none did better than chance at labeling incorrect actions, and next-step actions were correct only $\sim$52--70\% of the time--but they could produce remarkably human-like learning curves when trained as students with in-context learning.



 
\end{abstract}

\keywords{Intelligent Tutoring Systems \and 
         AI Benchmarks \and
         Simulated Learners \and
         Large Language Models
         }

\section{Introduction}

TutorGym is a system for interfacing several varieties of artificial intelligence (AI) agents---including large language models (LLMs), reinforcement learners (RL), and computational models of learning---directly with existing intelligent tutoring systems (ITSs) including CTAT tutors \cite{aleven2016example}, Apprentice Tutors \cite{smith2024apprentice}, and OATutors \cite{pardos2023oatutor}. TutorGym is inspired by Gym: an OpenAI system that enabled standardized comparisons between RL agents within several environments originally designed for human use, such as Atari games \cite{brockman2016openai}. While Gym focuses largely on games and only on RL agents, TutorGym extends this concept to education, providing a standard method for evaluating several kinds of AI agents on tasks relevant to learning engineering and the learning sciences. It can evaluate the tutoring performance of AI agents and their ability to simulate student learning within ITSs that have been used in real classroom studies. TutorGym fills major gaps in prior benchmarks by aligning AI assessments with research-backed adaptive tutoring approaches and the data collected when applying those approaches in the classroom.



LLMs' ability to provide step-by-step solutions to academic tasks has ignited interest in using them as automated tutors \cite{stamper2024enhancing}. No doubt, LLM capabilities for generating on-demand step-by-step solutions, explanations, and other content have compelling applications in automated tutoring. Nonetheless, blind enthusiasm for LLMs often understates the considerable new problems they introduce when applied to tutoring. LLM inference can be costly and consistently generates ``hallucinations''---erroneous outputs that often sound plausible but are incorrect, logically inconsistent, or completely fabricated. Large-scale benchmarks of mathematical tasks like GSM8K \cite{cobbe2021training} and the MATH dataset \cite{hendrycks2021measuring} measure LLM performance in generating final answers to math problems. High accuracies on these benchmarks (around 80-90\%) mask poor real-world performance. For instance, the GSM-symbolic benchmark altered problems from GSM8K by replacing constants (e.g., names and numbers) and adding distractor clauses \cite{mirzadeh2024gsm}. This substantially reduced LLM accuracy, suggesting that LLMs rely upon memorizing solutions instead of general-purpose problem-solving. 

Instead of only evaluating correct final answer generation, TutorGym comprehensively evaluates AI agents on step-by-step problem-solving and step-by-step solution recognition within real ITSs interfaces. TutorGym's ITSs teach a variety of topics, and each permits a variety of solutions. Many of these have been meticulously refined through iterative deployment and redesign to provide precise adaptive tutoring that supports the practice of fine-grained skills. 

\subsection{Intelligent Tutoring Systems}

\begin{figure*}[t!]
    \centering
    \includegraphics[width=.95\linewidth]{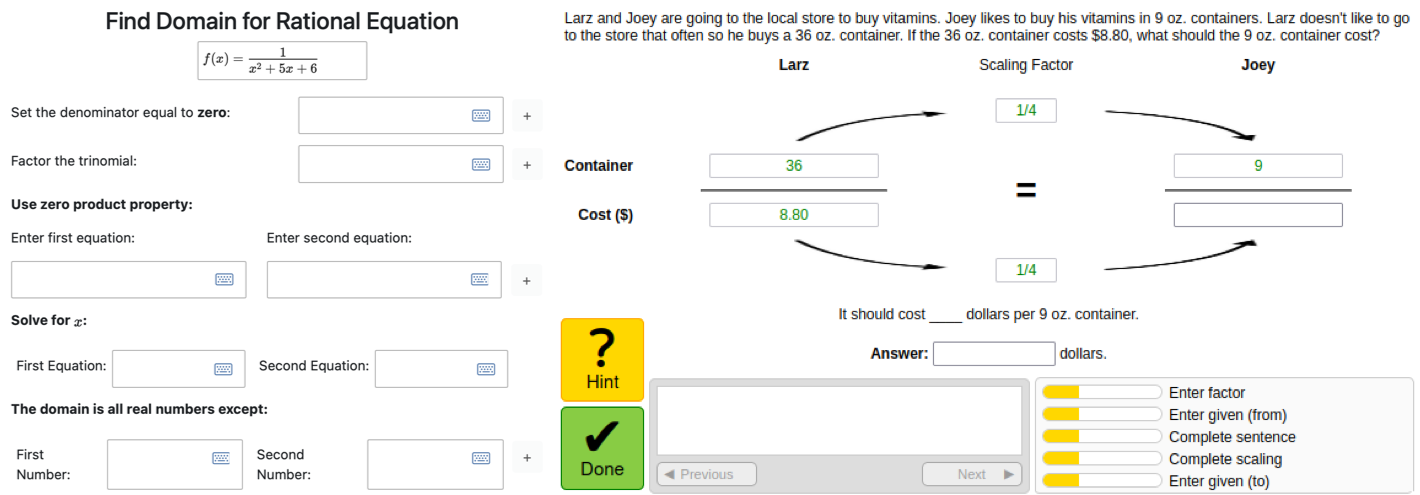}
    \caption{An Apprentice Tutor (left) and a CTAT tutor (right).}
    \label{fig:tutors}
\end{figure*}

ITSs are typically composed of at least three parts: 1) a tutor interface, 2) a tutor model, and 3) a student model \cite{nwana1990intelligent}. Careful interface design is a core element of many historical ITS successes. These interfaces present students with {\it scaffolding}, where the cognitive problem-solving steps are distributed across several interface elements---boxes to fill, buttons to press, widgets to interact with---that each test particular well-defined skills as part of a step-by-step solution strategy (see Figure \ref{fig:tutors}). This approach can be considerably more effective than simple non-adaptive assessments like online quizzes because the ITS can catch and address students' mistakes and lapses in knowledge at the step level instead of only at the end of full problem-solving attempts \cite{vanlehn2006behavior}. The tutor model tracks students' step-by-step answers and aligns their problem-solving progress with a number of acceptable strategies. There are several varieties of tutor models. Constraint-based models characterize correct solutions by the satisfaction of several independent requirements, that when violated, trigger highly-specific feedback \cite{mitrovic2010modeling}, conversational agents engage students in interactive dialogues \cite{graesser2004autotutor} and approaches like example-tracing and model-tracing track students solutions step-by-step to deliver correctness feedback, hints, and other forms of support that are aligned with particular problem steps, solution strategies, and common mistakes \cite{anderson1990cognitive}. In all of these cases, the role of the tutor model is to trace student problem-solving progress and adaptively align tutoring supports, such as on-demand hints, explanation messages, and step-by-step correctness feedback. 
Finally, the student model estimates students' mastery of individual knowledge components (facts, skills, or principles), to inform the selection of next practice problems and, in some cases, the behavior of the tutor model. 

\subsection{Evaluating AI Tutors with TutorGym}
LLM-based tutoring is a tantalizing modern challenge in learning engineering that envisions LLMs as adaptive support tools instead of just---as is all too often the case---tools for cheating that bypass learning entirely. Prior work has addressed issues of LLM inaccuracy by limiting their use to offline content generation \cite{calo2024towards} or applying constraints to their on-demand generation \cite{pardos2024chatgpt}. Others have developed ways of using LLMs to translate static text resources into more constrained formats similar to those used by traditional tutoring paradigms \cite{schmucker2024ruffle} like AutoTutor chatbots \cite{pal2024autotutor}. Efforts like these rein in LLM inaccuracy. Yet far too few efforts provide a substantive comparison between LLM tutoring and the decades of research-based automated tutoring that have preceded it. In some cases, LLMs have been evaluated against human-generated hints \cite{pardos2024chatgpt}, and group classroom instruction \cite{kestin2024ai}. Yet, these efforts rarely justify how LLMs might innovate on the high standard set by prior work.

ITSs have been shown to be as effective as, and in some cases more effective than, human tutors \cite{vanlehn2011relative,kulik2016effectiveness}. Their main drawback is the labor intensity of authoring them, which can require as much as 200-300 hours of development per hour of instruction, and typically begins with investigations of expert knowledge and student misconceptions using a method called Cognitive Task Analysis (CTA) \cite{aleven2016example}. ITS designs are also often refined through analyses of student data \cite{liu2017closing}. These empirical elements of ITS development engage with the notion that there is no ``magic bullet'' to learning or tutoring \cite{anderson2013implications,roll2014not}. Every capability requires mastery of several independent knowledge components \cite{koedinger2012knowledge}, and effective tutoring requires precisely assessing and supporting mastery of these elements. 

One may reasonably wonder if LLMs pose risks to learning engineers similar to those they pose to students. One cannot help but marvel when an LLM composes a step-by-step, textbook-like explanation before our eyes. However, on-demand assistance is not necessarily ideal or effective. Research-backed ITS design demands precision of understanding on the part of the learning engineer and precision of execution on the part of the ITS. It would be na\"{i}ve to expect that the fruits of this understanding could be superseded by simply regenerating content from the non-adaptive datasets used to train LLMs. Yet perhaps the best of both is possible. Future AI research may deliver empirically backed cognitive supports, with the convenience of LLMs and the precision of traditional ITSs. 

By interfacing various kinds of AI agents with a large repository of ITSs, TutorGym enables benchmarks of how precisely AI agents can deliver interface-based ITS instruction. Imagine a learning engineer who has done their homework: performed CTAs and designed a tutoring experience with a scaffolded interface. They may wish to skip the tedium of programming a tutor model and instead prompt an LLM to serve as a tutor model. TutorGym allows us to benchmark the feasibility of this scenario. Unlike the problem-solution pairs and sequential problem-solving that most LLMs are trained on, TutorGym environments deliver flexible adaptive tutoring. Many TutorGym tutors can support multiple solution strategies, allow for flexibility in step solution format and ordering, and utilize interface elements like buttons, checkboxes, and specialized widgets that deviate from purely text-based inputs. When students apply different sequences of actions in an ITS, they may end up in a wide variety of states that branch out from the initial problem presentation. For each of those states, each ITS in TutorGym can deliver step-specific hints and flexibly recognize a variety of possible correct next actions. In this work, we evaluate LLMs against tutor models in a large collection of intermediate tutor states. Specifically, in each state, we test if the AI can identify correct and incorrect next actions drawn from student data and demonstrate correct next actions.  

\subsection{Evaluating Simulated Learners with TutorGym}

Artificial Intelligence in Education (AIED) researchers have long sought to simulate human learning for scientific and practical applications in education. John Self, the AIED Society founder, promoted the idea of ``executable student models'' that learn directly from tutors and operate as ``wind tunnels'' for comparing and investigating ITS designs \cite{self1995computational,weitekamp2023computational}. Early simulated learner---like Sierra \cite{vanlehn1987learning,vanlehn1990mind} and Cascade \cite{vanlehn1999rule}---furthered these ideas, envisioning new applications where teachers practice tutoring with simulated learners \cite{vanlehn1994applications}, or students engage with them as learning companions \cite{ur1995steps,matsuda2015teaching}. More recent simulated learners---including SimStudent \cite{koedinger2015methods}, Apprentice Learner \cite{maclellan2020domain,weitekamp2020CHI}, and AI2T \cite{weitekamp2024ai2t}---can learn within tutor interfaces, inducing tutor models from minutes of interactive instruction. In line with Self's vision, several simulated learners have predicted experimental results \cite{rachatasumrit2023content,maclellan2025model} and reproduced student learning curves without fitting to student data \cite{maclellan2016apprentice,maclellan2023evaluating,weitekamp2019toward}. 

A large body of work has replicated student behaviors and numerical changes in student performance without simulating learning directly \cite{kaser2024simulated,radmehr2024towards}. LLMs have made it almost trivial to generate naturalistic human-like responses, meaning an undiscerning observer can be fooled by an LLM pretending to be a student. This has become a convenience in applications where the student ``simulation'' need only play the role of student, and not necessarily simulate learning. 

Evaluating simulated learners---those that actually learn---involves comparing their evolving behavior to the responses of human students. By enabling AI agents to work directly within ITSs used in classroom studies, TutorGym expands upon the set of publicly available environments where this comparison can be made. A common way to evaluate simulated learners is to compare their learning curves to those of human learners. This is typically framed as a near-zero-parameter comparison: aside from small adjustments for student-specific individualization \cite{maclellan2023evaluating,weitekamp2019toward}, models are not fit to student data, yet their \textit{a priori} predictions must align with it \cite{maclellan2025model,weitekamp2020investigating}. These strict evaluations promote the view of simulated learners as theory-first computational models of learning \cite{maclellan2016apprentice} where the mechanisms of learning are theorized, implemented, and executed on learning materials, and only then compared to student data in a final evaluation \cite{weitekamp2023computational}. 
Remarkably, computational models of learning have succeeded in closely aligning with human learning curves because they can learn from a few training experiences instead of the thousands to millions required by data-driven methods like reinforcement learning \cite{maclellan2025model,maclellan2021EDM,weitekamp2025decomposed}.  

LLM-based simulated learners have not been directly compared with computational models of learning on learning curve reproduction; however, TutorGym provides the means of making such a comparison on a large scale. As an initial example of how this kind of comparison might look, we compare the learning curves of LLMs working in TutorGym environments with real student learning curves. In this evaluation, the LLMs "learn" using in-context learning \cite{dong2022survey}: their next-step answer predictions are graded as correct or incorrect, and these experiences accumulate within their context window. 

\section{The TutorGym API}

TutorGym provides a standard Python interface for AI agents to interact with Cognitive Tutors (CTAT), Apprentice Tutors, and OATutors, and offers a common standard of interaction to unite these three distinct tutor platforms.

\subsection{Tutor Paradigms}
The Cognitive Tutor Authoring Tools (CTAT)\cite{aleven2006cognitive}, is the oldest of these tutoring platforms, and the most comprehensive in terms of domain diversity and flexible adaptive tutoring. CTAT interfaces include both standard HTML elements like text boxes, check boxes, and buttons; plus several specialized widgets like number lines and drag n' drop interfaces. CTAT has evolved through implementations in Java, Flash, and HTML. TutorGym supports CTAT's HTML-based tutors authored with example-tracing \cite{aleven2016example}: a programming-by-demonstration approach for authoring behavior graph-based tutor models. CTAT behavior graphs are essentially finite state machines with state nodes and action edges that can be augmented with additional functionality like unordered groups of actions, skippable actions, tutor-performed actions that trigger interface changes, and short programs that flexibly match variations in acceptable student answers. 

\begin{table}[t!]
    \setlength{\tabcolsep}{0.5em} 
    \renewcommand{\arraystretch}{1}
    \centering
    \caption{Difference in functionality between tutor paradigms.}
    \begin{tabular}{|c|c|c|c|}
    \hline
        \makecell[c]{\bf Tutoring Features} &                   \makecell[c]{ {\bf CTAT} \\ {\bf Example-Tracing}}  & \makecell[c]{{\bf Apprentice} \\ {\bf Tutors}} & {\bf OATutor}\\

        \hline
        \hline
        Flexible Answers           & yes     & yes     & yes \\
        Unordered Actions          & yes     & partial & no \\
        Optional Steps             & yes     & yes     & partial\\
        Tutor Performed Actions    & yes     & no      & no\\
        Contextual Control Flow    & partial & partial & no\\
        Accept Multiple Strategies & yes     & no      & no\\
        Adaptive Scaffolding       & partial & yes     & partial\\
        Adapt to Open-Ended Inputs & yes     & yes     & no\\
        Tutor Any Problem Instance & partial & yes     & partial\\
        Knowledge Tracing & several & BKT     & BKT \\
    \hline
    \end{tabular}
    \label{tab:tutor_comp}
\end{table}

Table \ref{tab:tutor_comp} outlines how CTAT example-tracing, Apprentice Tutors, and OATutors compare on several dimensions of functionality. For instance, Apprentice tutors are implemented with hierarchical rules that divide tasks into scaffolded subtasks \cite{siddiquihtn,smith2024apprentice}. These rules are implemented with a Python-based rule engine that enables flexible control-flow and grading behavior. Symbolic algebra tools like SymPy can be used with these hierarchical rules to relax the formatting requirements of mathematical answers. In principle, rule-based tutoring models like Apprentice Tutors enable more complex behaviors than CTAT example-tracing. For instance, rule-based tutors can capture solution paths that vary considerably between problem instances, such as when tutors require students to engage in contextual decision-making and decide between different strategies. Nonetheless, the Apprentice Tutors included in TutorGym at the time of writing this paper consist of purely sequential steps. However, the granularity of those steps can be controlled dynamically as part of next-problem selection by selecting different ``scaffold levels.''

The within-problem control flow complexity of CTAT's tutors are considerably greater by comparison: they sometimes permit dozens of correct next actions, while the Apprentice tutors permit just one or two. Both of these paradigms include domains where open-ended student inputs affect the acceptable answers in later steps. All three tutor paradigms have tools for tutoring arbitrary problem instances. Apprentice Tutors can randomly generate problems and tutor them on demand, while OATutor and CTAT example-tracing provide tools for mass-producing problems by replacing placeholders in templates.  

In terms of control flow complexity, OATutors are relatively simple by comparison to CTAT example-tracing and Apprentice Tutors. The OATutor problem pool consists mostly of one-step problems with some exceptions. However, short sequential ``tutor pathways'' offer a mixture of hints and extra scaffold steps that are available to students on demand \cite{pardos2023oatutor}. OATutors permit flexible recognition of mathematical expressions, similar to Apprentice Tutors.

Adding new problems to TutorGym is as easy as copying standard tutor models from its three supported paradigms. At the time of writing, TutorGym includes 10 domains from CTAT's Mathtutor suite with 732 problems (and more on the way), 30 Apprentice Tutor domains with random problem generators, and 188 OATutor domains with 5161 problem instances. In addition, two miscellaneous hard-coded domains: fraction arithmetic and multi-column addition with random problem generators, are included from a limited early version of TutorGym \cite{maclellan2021EDM}. These domains are implemented by programmatically generating finite-state machines using the same backend as our CTAT behavior graph interpreter. These tools are designed to be extensible to facilitate the integration of TutorGym with additional tutor paradigms in the future.

\subsection{Tutor-AI Interfaces}

\subsubsection{Symbolic Interface}

\begin{figure}[b!]
    \centering\includegraphics[width=.95\linewidth]{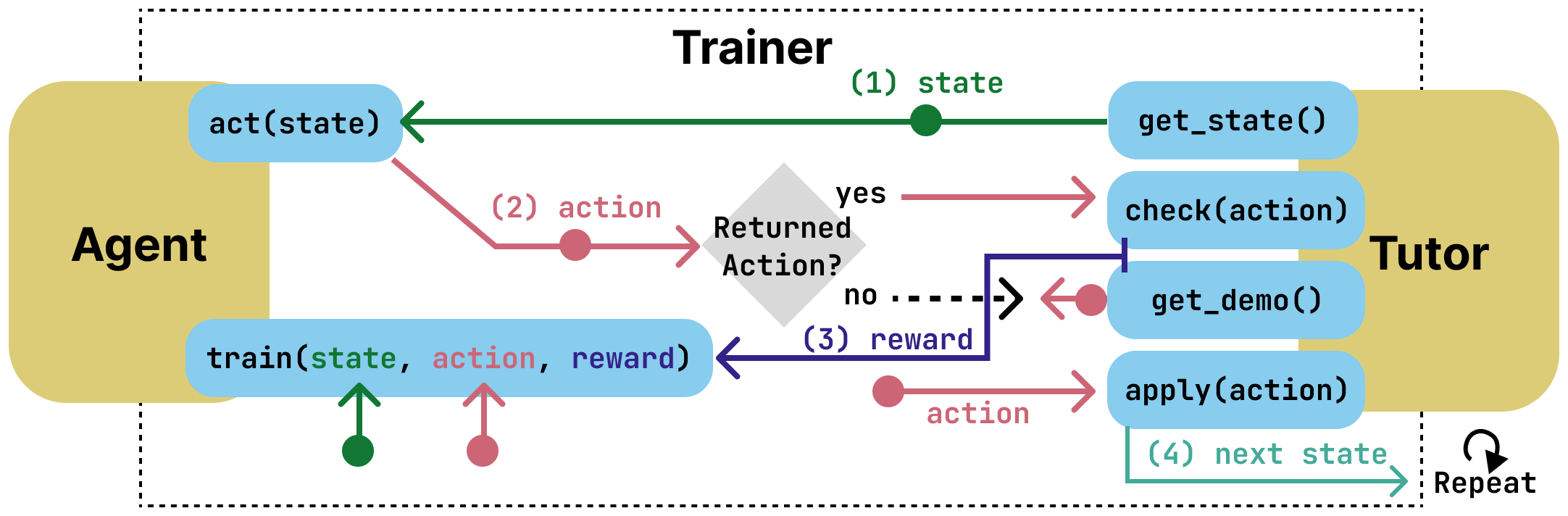}
    \caption{The \code{Trainer} class training an AI agent with a tutor.}
    \label{fig:trainer}
\end{figure}

TutorGym can adaptively tutor AI agents that implement two endpoints: \code{act(state)} and \code{train(state, action, reward)}. The \code{get\_state()} endpoint exposes each tutor's current \code{state} as a \code{ProblemState} object containing a JSON-like representation of the tutor's current HTML interface. When a \code{state} is passed to the agent's \code{act(state)} endpoint, it returns the next \code{action} it will attempt in the tutor interface. Each \code{action} instance contains a \code{(selection, action\_type, input)} triple (an SAI) that indicates 1) the \code{selection}: the \code{id} of the interface element being acted on, 2) a string indicating what kind of action is being applied, and 3) the content of the action (usually a string). For instance, an SAI for an action to fill a text field with the value ``7'' might look like \code{("field1", "UpdateTextField", "7")}. 

A \code{Trainer} (see Figure \ref{fig:trainer}) class mediates the interactions between an agent and tutor, over a sequence of training problems. In each state the \code{Trainer} queries the agent's \code{act(state)} function and evaluates it with the tutor's \code{check(state, action)} function to return a numerical \code{reward} (1 for correct, -1 for incorrect), that is passed to \code{train(state, action, reward)} to train the agent. If the agent provides no next action because it does not know what to do next then the \code{Trainer} calls \code{get\_demo(state)} in the tutor to instead demonstrate a `bottom-out' hint to the agent, that is passed to \code{train(state, action, reward)} in the form of a worked example \code{action} with a \code{reward} of 1. Optionally, the \code{Trainer} can force a demo \code{action} after a specified number of incorrect actions---a helpful feature for forcing training to move on when stubborn agents repeatedly guess incorrectly. When a correct action is produced, the \code{apply(action)} function applies the action to the tutor, yielding a next state.
Customizable logger classes can be provided to the \code{Trainer} to record each transaction. For instance, the \code{DataShopLogger} class records the standard transaction format used by PSLC DataShop \cite{koedinger2010data}. 

The symbolic interface can be used directly by Python-based computational models of learning like the Apprentice Learner \cite{maclellan2020domain} and AI2T \cite{weitekamp2024ai2t}, and serves as the foundation for the LLM and reinforcement-learning interfaces.

\vspace{-1.1em}
\subsubsection{LLM Interface}
The LLM interface is an agent wrapper that translates the \code{state}, \code{reward}, and next \code{action} demos from the symbolic interface into a prompt. By default the LLM is given the tutor state in JSON format. TutorGym can also open tutors in the browser and capture rendered images to augment the JSON-based state. The additional content of the LLM prompt is open-ended and up to researchers to design. In this work, we report an agent that implements \code{act(state)} by adding in-context examples collected from previous training. 

\vspace{-1.1em}
\subsubsection{Reinforcement Learning Interface}
TutorGym provides an \code{RL\_Wrapper} class that can be applied to any tutor, and provides fixed state and action spaces that can be used as an OpenAI Gym environment for training reinforcement learners. This interface simply maps values in the state, and SAIs in actions to fixed integer values that can be used as action indicators and indices for one-hot state vectors that can be read by neural networks. Prior work \cite{maclellan2021EDM,weitekamp2025decomposed} has shown that RL agents need tens of thousands of examples to succeed at tutoring tasks, and so in practice, only tutors that can randomly generate problem sets are compatible with the \code{RL\_Wrapper}.   

\vspace{-1.1em}
\subsubsection{Tutoring Evaluation Support: Completeness Profiles}
Evaluating agents on open-ended tutoring requires a comprehensive assessment of behaviors beyond step-by-step problem-solving. For this purpose, TutorGym can automatically generate \textit{completeness profiles}: collections of reachable ITS states with every correct next action produced by the tutor's \code{get\_all\_demos()} function. When student data is available, these profiles also include correct and incorrect student actions. Completeness profiles are built either by sampling several solution pathways in each problem of a large problem set or by replaying transactions from data to reproduce the tutor states experienced by students. An agent with 100\% correct tutoring behavior should be able to accurately grade the completeness profile---labeling the correct actions as correct and the incorrect actions as incorrect---and generate at least one correct action as a demo for each state. 

\section{Tutoring Evaluation}

\subsection{Method}

As an initial demonstration of using TutorGym to evaluate AI agents as interactive tutors, we tested several different LLMs, including Sonnet-3.5, Haiku-3.5, GPT-4o, and DeepSeek-v2.5 in TutorGym's 223 different tutor domains.\footnote{The snapshots were: \code{claude-3-5-sonnet-20241022}, \code{claude-3-5-haiku-20241022}, \code{gpt-4o-2024-08-06}, and \code{deepseek-v2.5:236b}.} These commercial models offered the best performance vs. cost tradeoff at the time of writing. The chosen open-source models were the most performant models runnable with Ollama on our server with four A40 GPUs. For Apprentice Tutors we generated completeness profiles from a student dataset containing about 15 thousand transactions. These profiles include both the correct next actions from each tutor's \code{get\_all\_demos()} output and the correct and incorrect actions taken by students. We were unable to acquire public student data for the CTAT Mathtutors and OATutors. In these domains, the completeness profiles are generated by sampling tutoring paths from all of the available problems in each of the 10 CTAT tutors domains and 5 problems per tutor domain for the OATutors. We augmented all profiles with incorrect actions produced by DeepSeek-v2.5 (which we could run locally).  


Recall that this evaluation simulates a scenario where a learning engineer prompts the LLM to operate as the ITS expert model in place of authoring a hard-coded one. To support this, we created specialized prompts that describe each tutor action with in-context examples and ask the agent to either generate a correct \code{action} for a given \code{state} (i.e., a worked example), or evaluate a given \code{state}--\code{action} pair by responding with ``yes'' if the action is correct or ``no'' if it is not.\footnote{see \textcolor{blue}{\href{https://github.com/Teachable-AI-Lab/tutor\_gym/blob/v1.0.0/tutorgym/eval/prompts.yaml}{tutorgym/eval/prompts.yaml}} and \textcolor{blue}{\href{https://github.com/Teachable-AI-Lab/tutor\_gym/blob/v1.0.0/tutorgym/eval/action\_type\_examples.yaml}{tutorgym/eval/action\_type\_examples.yaml}}}
OATutors and Apprentice tutors have well-defined steps that are fairly clear just from the labels and field ids in the problem state. For CTAT tutors, we supplement the prompt with domain-specific descriptions that explain their more complex control flows.\footnote{see \textcolor{blue}{\href{https://github.com/Teachable-AI-Lab/tutor\_gym/blob/v1.0.0/tutorgym/eval/ctat\_prompts.yaml}{tutorgym/eval/ctat\_prompts.yaml}}}

\subsection{Results}

Table \ref{tab:tut_eval} shows the results of our evaluation for all tutors and LLMs. All four of the LLMs that we tested are quite poor or heavily biased when grading step-by-step action correctness. The models with the highest accuracy at identifying correct actions were the worst at identifying incorrect actions, and no model exceeded chance (i.e. 50\%) at identifying incorrect actions. All models show poor accuracy---around 52-70\%---at producing next-step demonstrations. These accuracies are starkly lower than many published final-answer benchmark results. Consistent with other work \cite{gupta2025beyond}, they illustrate that current LLMs are impractical for providing out-of-the-box ``complete'' tutoring system support that grades and generates next actions in arbitrary intermediate tutor states.


\begin{table}[t]
    \centering
    \setlength{\tabcolsep}{0.5em} 
    \renewcommand{\arraystretch}{1}
    \caption{LLM accuracy at labeling correct and incorrect student actions and generating next step demonstrations (bottom out hints) across the tutors.}
    \begin{tabular}{|c|c|c|c|c|}
        \hline
        {\bf Tutor Platform} & {\bf LLM Model} & \makecell{{\bf Correct} \\ {\bf Accuracy}}  & \makecell{{\bf Incorrect} \\ {\bf Accuracy}}  & \makecell{{\bf Demo} \\ {\bf Accuracy}} \\
        \hline
        \hline
        \multirow{4}{*}{\makecell{CTAT Mathtutors \\ (10 domains)}}
        & Sonnet-3.5 & 61.92\% & 36.21\% & {\bf 56.50\%}
        \\
        & Haiku-3.5 & {\bf 81.06\%} & 25.05\% & 36.49\% 
        \\
        & GPT-4o & 28.11\% & {\bf 42.63\%} & 38.89\% 
        \\
        & DeepSeek-v2.5 & 59.96\% & 30.84\% & 39.33\%
        \\
        \hline
        
        \multirow{4}{*}{\makecell{Apprentice \\ (30 domains)}}
        & Sonnet-3.5 & 86.54\% & 46.56\% & 64.20\% 
        \\
        & Haiku-3.5 & {\bf 88.36\%} & 22.34\% & 49.73\% 
        \\
        & GPT-4o & 74.61\% & {\bf 49.80\%} & {\bf 70.75\%} 
        \\
        & DeepSeek-v2.5 & 82.00\%  & 48.92\% & 58.35\% 
        \\
        \hline
        
        \multirow{4}{*}{\makecell{OATutor \\ (183 domains)}}
        & Sonnet-3.5 & 79.59\% & {\bf 38.85\%} & {\bf 52.10\%}
        \\
        & Haiku-3.5 & {\bf 92.36\%} & 10.82\% & 36.63\% 
        \\
        & GPT-4o    & 71.20\% & 38.69\% & 51.07\%
        \\
        & DeepSeek-v2.5 & 90.87\% & 17.21\% & 43.89\% 
        \\
        
        \hline
    \end{tabular}
    \label{tab:tut_eval}
\end{table}

\subsection{Discussion}

The relatively poor performance of LLMs at these tutoring evaluations indicates a great deal of room for improvement in this space. We leave further experimentation with LLM tutoring in TutorGym to future work. However, we must caution that running these evaluations (including those in Section 4) cost more than \$730 US dollars in API fees to OpenAI and Anthropic, highlighting that inference costs are still a significant limitation to LLM-based tutoring approaches. We were able to run DeepSeek locally using available hardware.

\section{Simulated Learner Evaluation}

\subsection{Method}

We evaluate the Haiku-3.5 and GPT-4o LLMs as a simulated learner by running them through TutorGym's interactive training pipeline with the Apprentice Tutors. We selected these two models because they were the most cost-effective within their performance class. We omitted DeepSeek-v2.5 because inference speed was prohibitively slow on our available hardware. However, we hope to investigate the use of open-source LLMs as simulated learners in future work. 

To simulate tutored learning, we used a prompt similar to the one used for generating demos in the previous tutoring evaluation (see section 3.1).
The main difference is that when the agent receives feedback from the tutor, it stores the \code{state}, \code{action}, and \code{correctness} values and adds these as in-context examples to the prompt. The previously described tutoring evaluation prompt is essentially a zero-shot version of the prompt employed here.
As the LLM agent experiences more problems, it accumulates an increasing number of in-context examples and the prompt size grows. Given the high cost of large prompts and the difficulty of locally running open-source LLMs that support large contexts, we limited the prompt length to 50k characters (approx. 15k tokens using the GPT-4o tokenizer). To keep prompts to this size, we removed the oldest in-context examples whenever the length limit was exceeded.\footnote{We chose this limit to make it feasible to compare with open-source models on our available hardware in future work.} With this limit, the prompt could contain around 20-30 example states with their associated correct and incorrect actions, depending on the size of the states. The training sequence was not shuffled, and progressed one domain at a time, so previous training experiences from the same domain were usually retained within the context window.

To test this approach, we applied it to simulate learning with the Apprentice Tutors. We focused on this platform because we had access to human data for it, which made it possible to generate human learning curves. For each of the 30 domains, we trained the agents on the maximum number of problems solved by more than 4 of the 192 human learners in our data. This amounted to about 3 to 10 problem instances per domain. We ran five agent simulations with each LLM to generate simulated learning curves.

\begin{figure}[t!]
    \centering
    \includegraphics[width=.65\linewidth]{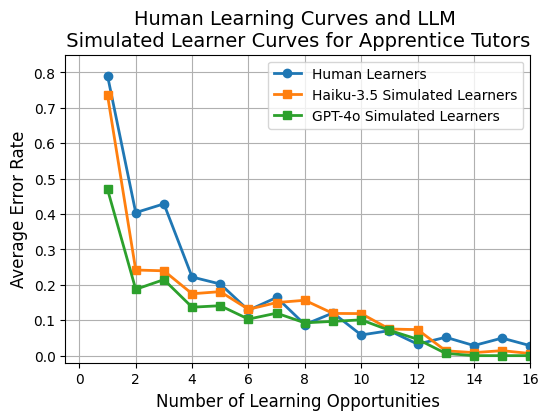}
     \caption{Learning curves from humans as well as Haiku-3.5 and GPT-4o agents with in-context learning solving problems in Apprentice Tutors.}
     \label{fig:learning_curves}
\end{figure}

\subsection{Results}

Figure \ref{fig:learning_curves} shows the human and LLM agent learning curves. These curves reflect the error rates on the first attempt applying each skill within each problem, following the convention used by PSLC DataShop \cite{koedinger2010data}. We associate each unique tutor interface element with a single skill and the curves average over all skills across the 30 tutors. Remarkably, the agents' error rates are qualitatively quite similar to the average human student error rate.

\subsection{Discussion}

It should be emphasized that although ``learning'' is in the phrase \textit{in-context learning}, it is arguably not a form of learning because it does not produce a permanent change in the model. The LLM's behavior is the product of a costly data-driven training process that includes more math problems than a human would solve in their lifetime.
The in-context examples guide the LLM in applying this knowledge within the tutor.
We hypothesize that they primarily provide information on the location and order of answer entry, rather than enabling the LLM to ``learn'' new math skills.

These results suggest that in-context learning may improve LLM tutoring capabilities over the zero-shot prompt we employed in section 3. However, we only evaluate whether the agent can generate a single correct action here, not its ability to grade complete action profiles.

For the purposes of illustrating how TutorGym can be used to facilitate the development of LLM-based simulated learners, this is a very interesting preliminary result. Before this demonstration, theory-driven computational models of learning \cite{weitekamp2023computational,maclellan2025model} have been the only variety of simulated learners to produce human-like learning curves within ITSs. Reinforcement learning agents, for instance, show orders of magnitude slower learning \cite{maclellan2021EDM}. LLMs, no doubt, have a great deal to offer in this space. We hope that TutorGym will become a valuable tool for AIED researchers to further these developments. 

\section{Conclusions \& Future Work}

TutorGym provides exciting new opportunities within AIED. LLM benchmarks typically lack the adaptive supports of ITSs, and prior tools for training simulated learners have been largely closed-source and limited to just a handful of domains. TutorGym offers tools for AI agents to interact with a growing suite of tutor domains (223 so far). We have illustrated how TutorGym can be used to evaluate AI agents as both tutors and as simulated learners. 

TutorGym has potentially many more applications than we have shown. For instance, TutorGym could be used to evaluate how effectively simulated learner-based authoring tools like AI2T \cite{weitekamp2024ai2t} are at learning ITS behavior from the author's instruction. In addition, researchers may use the hints available in its suite of tutors to evaluate automated hint generators. The possibilities are vast. TutorGym enables ITSs to be used not just as student-facing learning material, but also as environments for simulation and agent development. We look forward to seeing how others come to utilize it. 

\vspace{0.5em}
\noindent \small{{\bf Acknowledgments}
Funded in part by NSF \#2247790 and ARL \#W911NF2120101, \#W911NF2320203. Views are the author's and do not reflect those of the U.S. Govt.}

\bibliographystyle{splncs04}
\bibliography{references}

\end{document}